# Optimization of Waiting Time in H-R Coordination

Roy Someshwar
Dept. of Industrial Engineering and Management
Ben-Gurion University of Negev
Beersheba, Israel
Someshwar.Roy@eu4m.eu

Yoav Kerner
Dept. of Industrial Engineering and Management
Ben-Gurion University of Negev
Beersheba, Israel
kerneryo@bgu.ac.il

*Abstract—* **An analytical model of Human-Robot (H-R) coordination is presented for a Human-Robot system executing a collaborative task in which a high level of synchronization among the agents is desired. The influencing parameters and decision variables that affect the waiting time of the collaborating agents were analyzed. The performance of the model was evaluated based on the costs of the waiting times of each of the agents at the pre-defined spatial point of handover. The model was tested for two cases of dynamic H-R coordination scenarios. Results indicate that this analytical model can be used as a tool for designing an H-R system that optimizes the agent waiting time thereby increasing the joint-efficiency of the system and making coordination fluent and natural.**

*Keywords- Analytical Analysis; Optimization; Cooperative Systems; System Design; Cost function; Human-Robot Synchronization; Collaborative Manufacturing*

## I. INTRODUCTION

Human have a magnificent ability to collaborate with each other. This skill makes it possible to work in large teams and accomplish big tasks seemingly impossible for a single person. A fluent Human-Robot (H-R) system working as peers or co-workers can open new horizons in Robotics and it can completely change human life style and environment. Today this area is receiving considerable attention in the field of assistive robotics for personal care [1], [2], space robotics [3], [4] and social robotics [5], [6].

The working space of robot and human are usually strictly separated in industry. Off late, the significance and potential of Human-Robot Coordination (HRC) has also been realized in industrial robotics [7]. The current trend of the industrial sector is to develop smart, flexible, and easily customizable robots for diverse tasks involving close human-robot cooperation, sharing both work and time-space [8–10].

H-R Synchronization is defined as the harmonization between the actions of a human and a robot providing real-time coordination between them. A perfect synchronization is a case where the waiting times of the collaborating agents is zero at the point of handover. In other words, the transition is smooth and fluent.

H-R Synchronization, however, is a challenge that is prevalent in every type of H-R coordination system. This is due to the fact that the process of H-R Synchronization is influenced by several parameters related to the environment, the task and the agent [11]. A general framework for designing an H-R system that takes into consideration the given influencing parameters has been presented in [11].

The objective of the current paper is the *analytical analysis of the parameters influencing the waiting times or the unproductive times* of the collaborating agents in an H-R system. An optimization model of the H-R system is developed. It is implemented in two case-studies of dynamic H-R collaborative scenarios to study the behavior of the system when the cost of human delay varies with a certain range.

## II. METHODOLOGY

### A. Timing Control Model

In the manufacturing and assembly industries, the workers are often responsible for repetitive tasks. This led to the origin of the term 'takt time' which precisely meant time for one operation cycle per work station [12]. This term was initially coined by Toyota and later it became a common word in the car manufacturing industries.

The timing control model is based on the same principle. In this control model, a robot is controlled by a timer. The robot performs a series of pre-defined tasks at fixed intervals of time that is set by the end-user depending upon the needs and operational demands of the scenario. Sensors play no role in defining this cycle of operation or the action sequence. This however does not at all mean that there are no sensors in the robot. For example, if there is an obstacle in the work space of the robot then obviously the robot's safety system takes control of the situation and immediately stops.

In the timing control model, the operational cycle of robot actions is governed by only one parameter and that is time. The end-user has full control of this parameter and has the flexibility to define (feed or change) the time interval between consecutive events. One good example where this model suits the scenario is a pick and place robot in an assembly station.

This research was supported by the EU funded Initial Training Network (ITN) in the Marie-Curie People Programme (FP7): INTRO (INTeractive RObotics research network), grant agreement number: 23848TRO research project

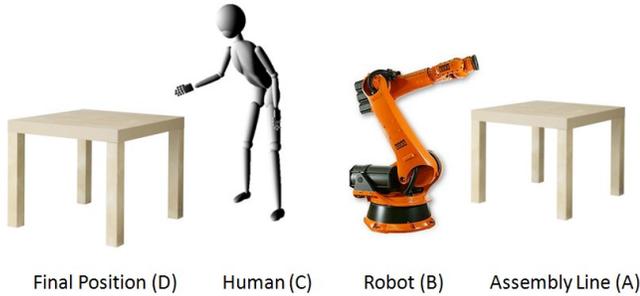

Fig 1: A Human-Robot Cooperative Scenario

*B. The Tested Scenario*

Fig. 1 gives an impression of the H-R system that has been analyzed in this research. The analyzed scenario consisted of a non-buffered, two-agent (human and robot) system executing a dynamic collaborative task that is repetitive (i.e., periodic) in nature with each agent responsible for an exclusive task. By exclusive task, we mean that each agent is responsible for an individual task that is independent of its partner and the coordination is required only at the pre-specified spatial point of handover at certain interval of time in every sequence of action.

From the right, the robot (B) picks up a metal block from an assembly line (A) and delivers it directly into the hands of the human (C). The human receives it and inspects the quality of the processed block and thereafter places it on another assembly line (D) or in the default section. The process continues repetitively from right to left over time.

This task was selected since it requires both speed and moderate accuracy which can be achieved by employing timing control model [13].

*C. Coordination Protocol*

In a Human-Robot system, it is important that the coordination mechanism between the agents is pre-defined with explicit protocol that explains the individual role of each agent in a collaborative task.

When timing control model is employed in an H-R system, there can possibly be two distinct protocols of coordination –

(i) Whoever comes first waits for the other

(ii) Robot never waits for the human but continues its periodic cycle. The human however, if arrives earlier waits for the robot.

In the first case, whoever arrive first waits for the other at the spatial point of handover until the handover is executed successfully. As a result, there is no cumulative error in this mode of coordination, and hence it is suitable for a robot that cannot be easily recalibrated.

In the second protocol, the robot never waits for the human at the point of handover but continues its cycle of periodic movement irrespective of the collaborating partner arriving or not arriving at the right time. The human, on the other hand, waits for the robot if it happens to arrive earlier. This means, if the handover is not successful in the first attempt, then the human waits for the second turn of the robot to repeat the same action. Such a protocol can be very useful for a scenario where a multi-tasking robot is employed that is also responsible for another job other than collaborating with the human. However, this mode of coordination may give rise to cumulative error, and hence recalibration of the system is necessary when a certain threshold level of waiting time is crossed to maintain the fluency of synchronization. Hence, the second protocol is suited for a robot which can be recalibrated easily.

*D. The Performance Measures*

An H-R collaborative task where the human and the robot are physically collaborating with each other requires the accurate anticipation of the spatial and temporal point of handover for an efficient synchronization of the process. *This paper deals with the analysis of the timing component of this handover to improve the fluency of synchronization*. We assume that another algorithm of the robot control system takes care perfectly of the spatial component of this mutual handover.

In such a scenario, if either of the collaborative agents is early or late in arriving at the pre-defined spatial point of handover, one will have to wait for the other. This time has been defined as the *waiting time* for the human and the idle/unproductive time for the robot. From the timing perspective, the goal of this analytical analysis is to study the nature and subsequently minimize the waiting times of each of the agents (human and/or robot) to improve the fluency of synchronization. The performance of our analytical model is thus evaluated by attaching a cost component to these idle/waiting times for each of the participating agents and then calculating this cost under various dynamic scenarios. Depending upon the needs and requirements of the dynamic scenarios, each of the agents is assigned with a different cost function and the behaviour of the H-R system is studied for various influencing parameters involved in the collaborative task.

*E. Human Delay Distribution*

The timing of human being is influenced by several intrinsic system variables. In other words, several psychological and neurological aspects of human are involved in a time-critical collaborative process including perceptual latency [14], temporal preparation [15], and rhythm of operation [16] [17]. The timing of human can also be influenced by several external factors including experience, fatigue and learning curve of the collaborating agent (human).

If the human is early or late at the point of handover, then this will give rise to an idle or unproductive time for the human. It can also be seen as a case of unsuccessful handovers or lack of fluency in the H-R system which has a direct effect on the overall productivity of the H-R system.

In the current analysis, the influence of human delay on Human-Robot synchronization and on the overall behavior of the system has been studied in detail. Based on the study, an optimization model has been suggested to minimize the waiting time / idle time of the agents.

### III. ANALYTICAL MODELING

In this section, the second coordination protocol has been analytically modeled, analyzed and studied in detail. As explained in the last section, in the second protocol, a robot never waits for the human but continues its periodic cycle. The human however, if arrives earlier waits for the robot.

In an analytical framework, this protocol can be modeled as a real-life scenario of a commuter and a public transport system. In order to analyze this problem, we follow the steps below:

i. Model the problem in analytical language for the given protocol thereby developing the problem statement.
ii. Taking into account the agent-intrinsic parameters affecting a human, we consider that the human has a random delay with (known) distributions
iii. The cost for human and robot in each periodic cycle is then identified and hence distinct cost function for each of the agent is developed
iv. This leads to the development of the objective function of the system.
v. The goal then is to find the agent (human and robot) timing that minimizes the expected cost (average cost).

#### A. The Analytical Model

Let us consider that the robot arrives at the point of handover at a regular interval of A seconds and it continues its periodic motion over and over again during the production cycle. We consider A as a decision variable. The optimized value of A is calculated according to the solution of the optimization problem that is developed later in Section C and solved in the form of case studies in section IV.

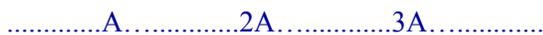

Fig 2: The periodic motion of Robot work cycle

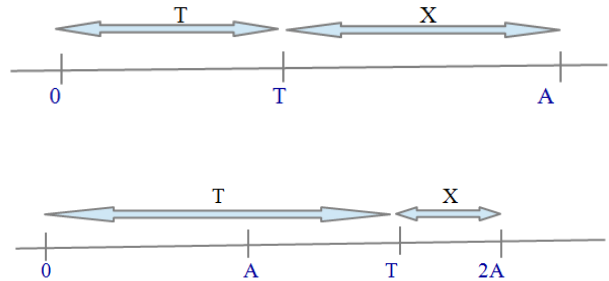

Fig 3: Illustration of the Scenario

In such a scenario, there can be two possibilities that have been well illustrated in the Fig. 3 above. The human arrival time T can be less than or greater than A. When it's less than A, the handover takes place in the first operational cycle of the robot. And when it's greater than A, it misses the first opportunity of handover and waits for the next robot cycle.

As mentioned earlier, let's say that the collaborating Human H has a random delay with (known) distribution. Human arrives after T seconds and waits for X seconds before the robot arrives. We consider T to be in the structure of T= t+Y, where 't' is the time that the human aims to arrive and Y is the random delay.

To better understand the nature of the human delay distribution, let's consider a real-life scenario of a commuter and a public transport system. Let's assume that the public transport is absolutely punctual and providing a service every A minutes. It is already known that the commuter has a general tendency to either delay and miss the transport or rush and reach the bus stop too early. Keeping in mind the commuter is not punctual, if we have to optimize one's time resources, the only way is to ask him/her to 'aim' an arrival time of 't' such that this 't' also takes care of the average delay/rush Y of the commuter.

#### B. The Decision Variable & the Influencing Parameters

In this analytical problem, we have two decision variables and three influencing parameters. The decision variables are t, the time when the human is scheduled and A, the time between two consecutive visits of the robot. The parameters are $C_R$, the cost for an unproductive visit of the robot, $C_H$, the human waiting cost per time unit and the function $F_Y(\cdot)$, the distribution of the human delay (while the random human delay is denoted by Y). For the sake of convenience, we define a random variable T=Y+t where T is the actual human arrival time.

The cost of waiting for the human W, is calculated as:

$$W = C_H [ R(t) - H(t) ]$$

$$W = C_H [A - T] \text{ for } A > T$$

where R(t) and H(t) are the times taken by the robot and human respectively to complete one round of operation. When A<T<2A, that is the human misses the robot in the first robot operational cycle, then two types of cost comes into play – the human waiting cost, W and the robot unproductivity cost, U.

In this case,

$$W = C_H [2A - T] \text{ for } A > T > 2A$$

$$U_{Robot} = C_R \text{ for } A > T > 2A$$

It is to be noted however, it is not necessary that the handover or the meeting between the agents would definitely take place at the latest in the second operational cycle. Depending upon the value of $F_Y(\cdot)$, the time T can vary from a value less than 't' and to a value far greater than A and hence it may take many unproductive robot operational cycles.

### C. The Objective Function of the H-R System

Given t and A, the decision variables, for any realization of T, the above equations of the waiting/unproductivity costs of human and robot respectively can be rewritten in a more generalized way as follows.

The number of robot's visits until the first productive one is $\left\lceil \frac{T}{A} \right\rceil$, where $\lceil a \rceil$ is the rounding up of a. Note that the cost of unproductivity per time unit is $\frac{C_R}{A}\left\lceil \frac{T}{A} \right\rceil$. The human waiting time, W is then $A\left\lceil \frac{T}{A} \right\rceil - T$. Thus, the total expected cost per time unit of the H-R system is:

$$TotalCost_{System} = Cost_{HumanWaiting} + Cost_{RobotWaiting} \text{ (eqn. 1)}$$

Hence, the *objective function of the H-R system* can be expressed in the following way:

$$Cost(t, A) = E_T\left[C_H\left(A\left\lceil \frac{T}{A} \right\rceil - T\right) + \frac{C_R}{A}\left\lceil \frac{T}{A} \right\rceil\right].$$

The solution of this optimization problem is to find t and A that minimizes $Cost(t, A)$.

As mentioned earlier, it is to be noted that although the random part of human time is referred to as "delay", this value part can also be negative. This is because the human is sent to the station with the aim to be there at time t, but the human might rush and arrive earlier than t. In terms of the delay's distribution function, we mean $F_Y(0)$ may be positive.

## IV. EXACT SOLUTION IN CASE STUDIES

In this section we exemplify the model using two case studies. In each of the case study a different example of delay distribution is used. In each case study we:

a) Derive the ingredients of the expected cost (human waiting and robot unproductivity).

b) Optimize the human 'aimed' time (considering the random delay) and the robot cycle time.

c) Study the behavior of the optimal cost and optimal cycle time in terms of the parameters.

### A. Case-Study I

Assume that $F_Y(y) = y$. That is, the delay is uniform along the interval (0,1). Clearly, for any choice of t, the optimal choice of A is in between t and t+1. Furthermore, if we choose t with a value larger than 1 it will be suboptimal. This is because, in that case either a needles human waiting or an unproductive robot visit is guaranteed (a unit of time is simply lost in every cycle). The expected human waiting time, given a pair $(t, A)$ (by trivial integration) equals

$$\frac{A - At + t^2}{2}.$$

The expected number of unproductive robot visit equals

$$\left\lceil \frac{1+t}{A} \right\rceil\left(1 + t - \frac{1}{2}\left(\left\lceil \frac{1+t}{A} \right\rceil - 1\right)\right) - t$$

The total cost of the H-R system is then

$$C_h \frac{A - At + t^2}{2} + C_r\left(\left\lceil \frac{1+t}{A} \right\rceil\left(1 + t - \frac{1}{2}\left(\left\lceil \frac{1+t}{A} \right\rceil - 1\right)\right) - t\right).$$

Note that this function has discontinuities (because a part of it has step functions) which makes the analysis a bit delicate. We next fix the robot unproductivity cost $C_r = 1$ and present the optimal cost of the system and optimal $(t, A)$ when the human cost $C_h$ ranges between 0.1 and 20. It is not surprising, due to the uniform nature of the human delay, that the optimal t equals 0 for any choice of the parameter. That is, the human should be aimed to arrive as soon as possible. The following figures show the optimal A and the optimal system cost as a function of $C_h$.

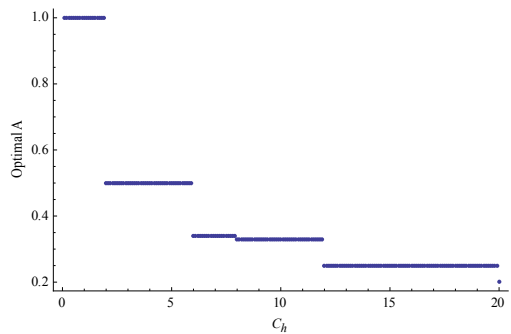

Fig 4: The graph shows the optimal value of A as a function of $C_h$

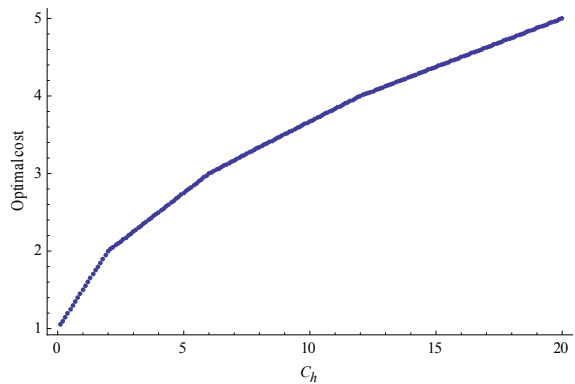

Fig 5: The graph shows the optimal cost of the system as a function of $C_h$

### B. Case-Study II

Now, let's study the behavior of the H-R system for a different delay distribution of the collaborating human. Let's assume that the human delay distribution is $F_Y(y) = 1 - e^{-y}$. That is, the delay follows an exponential distribution with rate 1. Due to the memory-less property of the delay distribution, it is suboptimal to choose a value of A smaller than t.

By simple integration, the expected human waiting time is calculated as:

$$A - (1+t) + \frac{Ae^t}{e^A - 1}.$$

The expected number of robot visits until the first successful handover is equal to

$$1 + \frac{e^t}{e^A - 1}.$$

We present the results for values of $C_h$ between 0.1 and 20, while fixing $C_r = 1$. The optimal value of 't' turns out to be always t=0. This is not surprising because practically aiming for arriving later than t=0 is simply a waste of time.

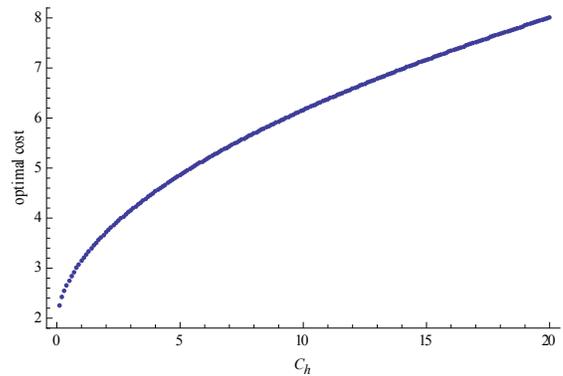

Fig 6: The optimal cost of system as a function of $C_h$

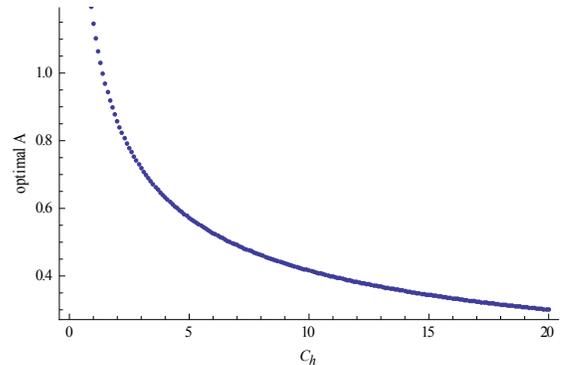

Fig 7: The optimal A as a function of $C_h$

## V. CONCLUSION AND FUTURE WORK

This paper shows an alternative approach to make human robot interaction fluent and natural. The analytical model takes into account the relevant context-dependent variables and hence it can optimize the waiting time or the unproductive time of the collaborating agents, thereby providing better fluency of handovers and possibly increasing the throughput of the H-R system.

A methodology for developing an optimization model of H-R system and its implementation in different case studies has been presented in this paper. This provides the research community with an excellent tool to study the design and behavior of an H-R coordination system.

It can be further developed to analyze an H-R system when the robot operation is controlled by an intelligent adaptive control model [18]. The case-studies can be further expanded for four different delay distributions of human namely experienced end-user, novice end-user, user with fatigue and user with high learning curve. The results of these case-studies will help in designing more human-centric H-R system. These analytical results can also be validated by implementing them in robot platforms and running experiments with human subjects on H-R synchronization.

The developed model and the general framework of the presented approach can also be used as an analytical method for analyzing human-robot systems to obtain better synchronization among them. The study is a first step towards the development of a comprehensive framework for the specification of human-robot coordination mechanisms.

ACKNOWLEDGMENT

The industrial robots from Yashkawa Motoman Robotics installed at the Paul Ivanier Center for Robotics Research and Production Management were used in carrying out this research. The authors also acknowledge the kind support of Prof. Joachim Meyer and Prof. Yael Edan, the Rabbi W. Gunther Plaut Chair in Manufacturing Engineering, Ben-Gurion University of the Negev.